\documentclass[runningheads]{extras/llncs}
\usepackage{graphicx}
\usepackage{todonotes}
\usepackage{booktabs}
\usepackage{algorithm} 
\usepackage{algpseudocode}
\usepackage{commath}
\usepackage{caption}
\usepackage{subfigure}
\usepackage{bm}
\usepackage{textcomp}
\usepackage{enumitem}

\begin{document}

\title{A Graph Convolutional Network Composition Framework for Semi-supervised Classification}

\titlerunning{GCN Composition Framework}

\author{Rahul Ragesh \and Sundararajan Sellamanickam \and Vijay Lingam \and Arun Iyer}

\institute{Microsoft Research, India}
\authorrunning{R. Ragesh et al.}

\maketitle              
\begin{abstract}
Graph convolutional networks (GCNs) have gained popularity due to high performance achievable on 
several downstream tasks including node classification. Several architectural variants of these 
networks have been proposed and investigated with experimental studies in the literature. 
Motivated by a recent work on simplifying GCNs, we study the problem of designing other variants and 
propose a framework to compose networks using building blocks of GCN. The framework offers flexibility to compose and evaluate different networks using feature and/or label propagation networks, linear or non-linear networks, with each composition having different computational complexity. We conduct a detailed experimental study on several benchmark datasets with many variants and present observations from our evaluation. Our empirical experimental results suggest that several newly composed variants are useful alternatives to consider because they are as competitive as, or better than the original GCN. 
\end{abstract}

\section{Introduction}
Semi-supervised learning (SSL) methods \cite{ssl_survey} are designed to address the problem of learning with both labeled and unlabeled examples. Developing powerful SSL methods have been of significant importance for nearly a couple of decades as: (a) getting labeled examples are expensive or difficult, and unlabeled examples are easily available in many applications and (b) achieving good generalization performance in label scarcity scenario is often a necessity with cost benefits. We are interested in one important class of SSL methods, Graph based SSL methods. These methods make use of additional relational information between entities that may be explicitly available (e.g., links between web-pages, citations) or derived from existing data (e.g., nearest neighbor graphs derived from features) \cite{lp_zhu,lp_zhu_zoubin_lafferty}. Methods and models used in problem formulation differ in how the available graph information is used (e.g., regularization, label propagation and feature propagation) in solving the underlying semi-supervised learning problem \cite{manifold_reg,lp_zhu,gcn}. 

In the last few years, there has been significant interest in designing various (a) graph neural networks (GNNs) and their variants \cite{survey_zhang_cui,survey_zhang_yin,survey_wu,survey_chen,survey_kinderkhedia} and (b) learning algorithms \cite{clustergcn,gcn_webscale} to train these networks for semi-supervised learning problems. Popular graph based neural networks include graph convolutional networks (GCN) \cite{gcn}, graph attention networks (GATs) \cite{gat}, graph recurrent networks \cite{gated_graph_sequence_networks}, etc. These networks combine graph structures with functionalities such as feature propagation (through several layers), transformation and nonlinear activation in several interesting ways to learn embeddings for nodes, edges and sub-graphs. Learned embeddings serve as key features for many downstream tasks (e.g., classification, clustering, ranking) that arise in many applications. GCN based algorithms achieve state of the art performance with significant improvement over traditional SSL methods \cite{gcn,gat,gam,planetoid}. Recently, there has been some progress made to simplify GCNs (SGCNs) \cite{sgcn}. The authors \cite{sgcn} demonstrated that significant performance improvement comes from \textit{feature propagation}, and nonlinear activation functions used across convolutional layers in GCN are not critical in several applications. Removing the nonlinear activation functions results in a linear model that still maintains the flavor of using multi-layer receptive field of convolution network. 

Motivated by the SGCN work \cite{sgcn}, we study the problem of designing other GCN variants. We look at the convolutional layer more closely and suggest an alternate way to split the layer into two components, smoothing and feed-forward layers. We propose a framework to compose graph convolutional network variants using these components. Note that these components are constructed using the building blocks of GCNs only. The notion of smoothing layer is generic and we use this layer to construct a propagation network (\textit{aka} a multi-layer smoothing network). Besides using this network for the purpose of feature propagation, we show how this can be used for propagating class probability distributions. One key point is that the propagation network makes use of the multi-layer GCN receptive field in both cases. Furthermore, when combined with the other component of feed-forward layer, the framework offers flexibility to compose network variants using different propagation choices along with linear or nonlinear transformations. 

We illustrate the use of our framework in constructing several variants in Figures 1-3 (Section 3). As we can see, SGCN is a special case of GCN composable from this framework. A couple of GCN variants show how to (a) use label propagation with multi-layer perceptron and (b) combine feature and label propagation in GCN. These variants have different computational complexities providing a trade-off between computational cost and performance. It is important to note that the composable variants are useful alternatives to experiment and validate on real-world applications, and compare along with the original GCN.   

Our contributions are the following: 
\begin{enumerate}
    \item We propose a framework to compose graph convolutional network variants using smoothing and feed-forward layers. The framework offers flexibility to compose and evaluate a variety of networks with different propagation and transformation capabilities, and computational complexities.
    \item Using the notion of smoothing layer, we suggest a simple way to propagate labels in GCNs and attempt to answer questions such as: \textit{is label propagation as good as feature propagation in GCNs?} 
    \item We conduct a detailed experimental study on several benchmark datasets with many GCN variants and our empirical experimental results suggest that several newly composed variants (e.g., MLP and GCN with label propagation) are as competitive as, or better than the original GCN. 
    \item We also present empirical observations from comparison of networks along different dimensions (e.g., feature propagation versus combined feature and label propagation, linear versus nonlinear).
\end{enumerate}

Our paper is structured as follows. In Section 2 we present preliminaries and background on GCNs. The proposed GCN composition framework and experiments are discussed in Sections 3 and 4. Related work (Section 5), Discussion and Future work (Section 6) are followed by Conclusion (Section 7).  
\section{Preliminaries and Background}
We consider node classification problem in graph based semi-supervised learning setting. We are given a graph $\mathcal{G}$ specified by an undirected adjacency matrix ($A$). Each node in the graph has a feature vector $\mathbf{x} \in {\mathcal R}^m$ and a label $y$ where $y$ belongs to one of $M$ classes. In addition, a set of labeled nodes (${\mathcal L}$) with their respective labels $({\mathcal Y}_L)$ and a set of unlabeled nodes $({\mathcal U})$ are given. Thus, given ${\mathcal D} = \{(x_i,y_i), i \in {\mathcal L}, \{x_j, j \in {\mathcal U}\}\}$ and $A$, the goal is to learn a network function to infer labels of unlabeled nodes (${\mathcal U}$).  

We add self-nodes and define an augmented adjacency matrix: ${\tilde A} = I + A$, as suggested in \cite{gcn}. Let ${\tilde D}$ denote the diagonal node degree matrix corresponding to ${\tilde A}$. In graph convolutional network \cite{gcn}, a symmetric normalized matrix ${\tilde S = {\tilde D}^{-\frac{1}{2}}\tilde A}{\tilde D}^{-\frac{1}{2}}$ is used for smoothing and propagating features across the layers. Here, smoothing refers to the basic operation of premultiplying feature vector by ${\tilde S}$, e.g., ${\bar {\textbf{x}}} = {\tilde S}\textbf{x}$. In our work, we use a row normalized version of ${\tilde S}$, call it $S$. This is because label smoothing involves aggregating node class probability vector of neighboring nodes and the resultant smoothed output should also be a probability vector.   

\subsubsection{Graph Convolutional Network} 
A graph convolutional network \cite{gcn} is a multi-layer network architecture with the first (\textit{aka} input) layer taking the node feature matrix ($X \in {\mathcal R}^{n \times m}$ where $n$ and $m$ denote the number of nodes and input feature dimension) as input. With $H(0) = X$, each layer consumes its previous layer's output (i.e., embeddings of all nodes in the graph, denoted as $H(k-1)$) and produces a new embedding matrix ($H(k) \in {\mathcal R}^{n \times d}$) where $d$ is the embedding dimension). Thus, the core convolutional layer is specified as: 
\begin{equation}
   H(k) = \sigma({\tilde S}H(k-1)W_k)
\label{gcnlayer}
\end{equation}
where $W_k$ and $\sigma$ denote $k^{th}$ layer weight matrix and nonlinear activation function (e.g., a rectilinear activation function $\text{ReLU}(\cdot)$). 

\subsubsection{Building Blocks of GCN Layer} \cite{sgcn} noted that the convolution layer (\ref{gcnlayer}) in GCN is composed of three blocks. They are: 
\begin{enumerate}[label=(\alph*)]
    \item \textit{Feature Smoothing}: ${\tilde H}(k) = {\tilde S}H(k-1)$, a matrix pre-multiplication operation that smooths input features or intermediate embedding output.
    \item \textit{Feature Transformation}: ${\tilde H}(k)W$, a post multiplication operation that transforms node embeddings (possibly, to a different dimension).
    \item \textit{Nonlinear Transition} function: $(\sigma(\cdot))$, a point-wise function that does nonlinear transformation. 
\end{enumerate} 

In the node classification problem, the final layer is a softmax logistic regression layer which takes smoothed ${\tilde H}(L) = {\tilde S}H(L-1)$ as input and outputs the class probability matrix $P$. That is, 
\begin{equation}
P(Y|X) = \text{Softmax}({\tilde H}(L)W_S)
\label{LRlayer}
\end{equation}
where $P(Y|X)$ denotes the node class probability matrix (of dimension $n \times M$) with $i^{th}$ row representing $i^{th}$ node's class probability vector and $W_S$ denote the softmax layer weight matrix $W_S$ (of dimension $d \times M$). 

\subsubsection{Feature Propagation} It is useful to note that the feature smoothing block is the one that differentiates (\ref{gcnlayer}) from a feed-forward network layer. It makes use of the underlying graph information (${\tilde A}$) and feature propagation happens through multiple layers with nonlinear transformation. \cite{sgcn} empirically showed that intermediate nonlinear transition function is not critical in several applications. When this function is dropped in (\ref{gcnlayer}) it is easy to see that feature propagation effectively becomes a sequence of matrix pre-multiplications, i.e., $H(L) = {\tilde S}^LX$ and the final embedding matrix ($H(L)$) can be pre-computed. Further, the model becomes linear along with the softmax logistic regression layer, resulting in a simple and efficient model making use of the multi-layer receptive field of GCN. 
\section{A Graph Convolutional Network Composition Framework}
In this section, we present a framework to compose different GCN variants. Motivated by \cite{sgcn}, the purpose is to define components that are useful to build simpler variants to experiment with and to validate on real-world datasets. Furthermore, it helps to validate hypotheses related to usefulness of different building blocks and their composition. In principle, we use the same building blocks described earlier. However, we present a different way to split or decouple (\ref{gcnlayer}) into two layers as follows:  
\begin{enumerate}[label=(\alph*)]
    \item Smoothing Layer: A smoothing layer takes an input matrix (${\tilde S})$ and uses the graph structure to smooth the input, i.e., ${\tilde H(k)} = \sigma({\tilde S} H(k-1))$. 
    \item Feed-forward Layer: This is the well-known nonlinear transformation layer used in feed-forward networks, i.e., $G(k) = \sigma(G(k-1)W_k)$.
\end{enumerate} 
Note that (\ref{gcnlayer}) is recovered by composing the smoothing layer with identity activation function and the feed-forward layer. Since the feed-forward layer is well-known, the only high level component that we introduce is the \textit{smoothing layer}. This layer is useful beyond feature smoothing and we use it to construct a network called \textit{propagation network} (i.e., \textit{multi-layer smoothing network}) by composing a sequence of smoothing layers similar to the feed-forward network. For example, we use the propagation network for label propagation (LP) as explained below. A propagation network is specified by the number of layers, one or more smoothing matrices and activation functions with hyper-parameters to tune. 

\begin{figure}
    \centering
    \begin{subfigure}
        \centering
        \caption{Building Blocks}
        \includegraphics[width=0.8\textwidth]{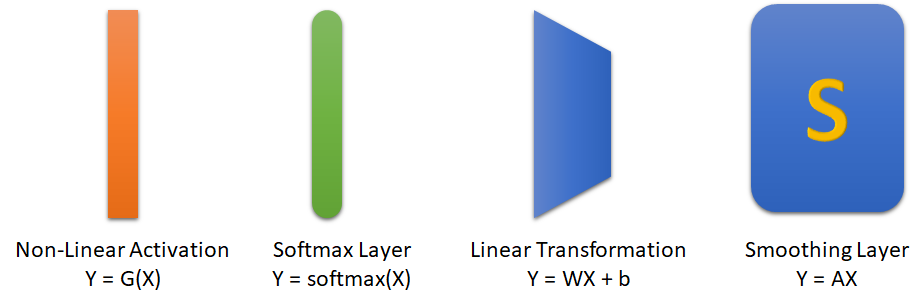}
    \end{subfigure}

    \begin{subfigure}
        \centering
        \caption{(a) MLP (b) MLP+LP}
        \includegraphics[width=0.4\textwidth]{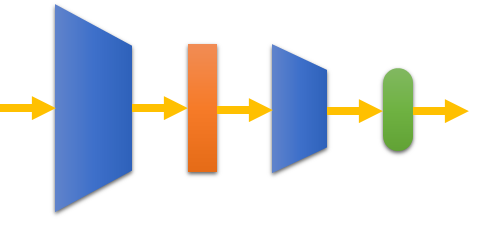}
        \qquad
        \includegraphics[width=0.4\textwidth]{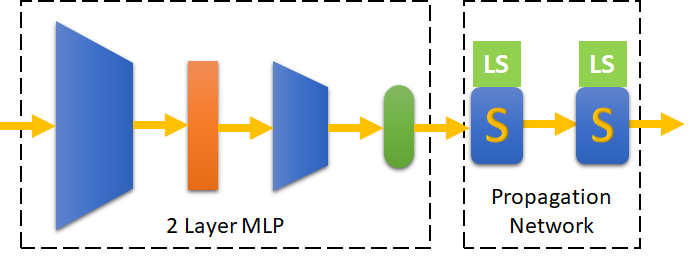}
        \label{fig:mlp_lp}%
    \end{subfigure}

    \begin{subfigure}
        \centering
        \caption{GCN+LP}
        \includegraphics[width=0.6\textwidth]{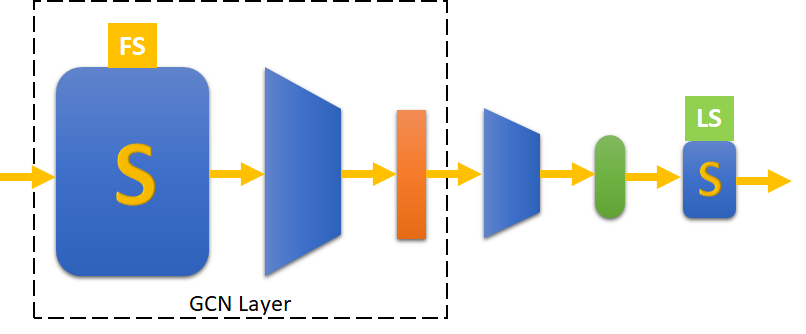}
    \end{subfigure}
    
\end{figure}

A natural question then is: \textit{what do we gain by defining these components?} It is true that any GCN variant composed using smoothing and feed-forward layers can be composed using the low level building blocks. However, there are several advantages to compose using high level components. It offers a different level of flexibility than the restriction imposed by using the convolutional layer (\ref{gcnlayer}) repeatedly. It helps to: (a) compose newer architectures with specialized functionalities such as propagation and non-linearity, which otherwise may not be obvious, and (b) to visualize the architecture better and (c) understand the usefulness of individual components along with trade-off between computational cost and performance.

\subsection{Examples of Composed Networks} 
We illustrate below how the framework can be used to compose different variants of GCNs.
\begin{enumerate}
    \item \textbf{Simplified GCN (SGCN):} SGCN is composed by a feature propagation (FP) network followed by the linear softmax logistic regression layer.
    \item \textbf{FP + MLP}: Though SGCN is a useful variant when linear models are sufficient, nonlinear models are helpful in some applications as our experimental results demonstrate. One simple way to do this is to compose feature propagation network with a multi-layer perceptron. As is the case in SGCN, features can be pre-computed with propagation network and only MLP training is needed. 
    \item \textbf{MLP + LP}: Another useful nonlinear model can be composed using multi-layer perceptron with class probability distribution output feeding to a label propagation network. 
    \item \textbf{GCN + LP}: One advantage of this variant is that it offers flexibility to experiment with reduced number of GCN layers and it compensates with simpler propagation networks. This is an example where feature and label propagation occur simultaneously. Since the label propagation layer does not have learnable weights, the number of model parameters is lesser in the combined network (assuming fixed number of layers).  
\end{enumerate}
Figures 2 and 3 depict a few example architectures. In the experiment section, we demonstrate the usefulness of a few variants. For example, our experimental results show that the MLP-LP variant is quite competitive to the original GCN and the performance of GCN can be improved when combined with LP in some datasets. 

\subsection{Label Propagation with Propagation Network} 
We propose a simple mechanism to propagate labels using the multi-layer receptive field structure with the propagation network. To propagate labels, the node probability matrix $P(Y|X)$ is passed as input to a smoothing layer to produce a smoothed output probability matrix, i.e., ${\bar P}(Y|X)$ and this process is repeated $L$ times where $L$ denotes the number of layers in the propagation network. As mentioned in Section 2, the pre-multiplier matrix ($S$) that encodes the graph structure is row-normalized to ensure valid probability outputs. In its simple form, a propagation network has a single smoothing matrix with identity activation function (i.e., $f(z) = z$) used in all the layers. In this case, label propagation results in ${\bar P}(Y|X) = S^{L}P(Y|X)$. The important aspect of using label propagation in the network is that the model parameters are learned by taking the propagation effect into account. This happens because class distributions of all neighboring nodes play a role while computing the gradient of the loss term for each labeled example. In the experiment section, we show that label propagation provides performance that is at least as good as feature propagation. 

Similar to observations made in \cite{sgcn} on $L$ with feature propagation, it is often sufficient to have a small number of layers (e.g., L = 2 or 3) to get competitive performance. The performance may degrade with label over-smoothing in some cases. In \cite{gcn,sgcn} experimental results using several propagation matrices were reported for feature propagation. We report similar experiments with label propagation. Furthermore, we parameterize using hyper-parameters and present results on a few benchmark datasets. 

\subsubsection{Feature Smoothing Versus Label Smoothing} 
We relate feature smoothing and label smoothing using a recent work \cite{gcn_lpa} which proposes an approach that unifies GCN and label propagation. The authors analyzed feature and label propagation by considering the basic smoothing operation. They derived a theoretical relationship showing feature smoothing ensures label smoothing within certain assumptions (e.g., node feature is a smooth version of its neighbors features). However, \cite{gcn_lpa} also points out some example cases where such assumptions may not hold in practice and suggests that this issue can be addressed by adapting the graph structure (i.e., learn $A$) using label propagation algorithm along with GCN model. In this work, our goal is not to adapt $A$ but to suggest label propagation as an alternative to feature propagation and to possibly combine the two. This seems to be a reasonable variant to experiment with and to validate on real-world datasets because assumptions do get violated in practical scenarios.     

\subsection{Learning Formulation}
We are interested in solving a node classification problem in a semi-supervised setting and the objective function is essentially the cross-entropy loss defined over the labeled examples \cite{gcn}. That is, 
\begin{equation}
    F(W) = - \sum_{i \in {\mathcal L}} \sum_{m=1}^M y_{i,m} \log {\bar p}(y_{i,m}|{\bar {\bf X}};W)
\label{gcnvariant}
\end{equation}
where $W$ denotes the learnable model weights and is dependent on whether a linear or multi-layer perceptron or a GCN model is used. We indicate presence of feature and label propagation networks by using the \textit{bar} notation. For example, the presence of feature propagation network at the input is indicated as: ${\bar X} = \text{FPN}(X;S,L)$. Similarly, the presence of label propagation at the output is indicated as: ${\bar P}(Y|X;W,S,L) = \text{LPN}(P(Y|X;W);S,L)$ where $S$ and $L$ denote the smoothing matrix and number of layers in the propagation network. 

\subsection{Computational Complexity} 
In Table 1 we give examples of several networks with \textit{function} and computation \textit{cost} details. We have included GCN and SGCN for reference. The examples cover linear and non-linear models combined with feature and label propagation either exclusively or jointly. For brevity, we provide \textit{function} details only for 2-layers scenario. On the other hand, the cost is given for the general case with $L$ layers (in GCN and MLP) and $L_l$ layers in the label propagation network. 

There are three major operations that incur computational costs and some of them are shared across networks.
The propagation network cost is dependent on whether propagation happens (a) at the input or intermediate layer and (b) with features or class probability scores. Networks that use propagated feature matrix at the input (e.g., SGCN and FP + MLP) incur one time pre-computation cost and are efficient during training. But, when propagation happens in intermediate layers (e.g., GCN), the cost is $O(LN_Ed)$ (assuming the feature dimension is fixed across the layers) where $L$, $N_E$ and $d$ denote the number of layers, edges and feature dimension respectively. On the other hand, the label propagation cost is $O(LN_EM)$ where $M$ is the number of classes and is cheaper compared to feature propagation when $M$ is small compared to $d$. The nonlinear transformation and Softmax(SMx) classifier layers cost $O(Lnd^2)$ and $O(ndM)$ respectively where $n$ denotes the number of nodes. 

SGCN is the most cost efficient network due to removal of nonlinear activation functions making pre-computation of propagated feature matrix (${\bar X}$) possible. The cost of other networks is dependent on the propagation types (i.e., feature and/or label), number of layers and presence/absence of MLPs. Thus, the framework offers flexibility to experiment with different networks subject to cost constraints, if any.    

\begin{table}[]
\centering
\caption{Composed Networks with Computational Cost. The \textit{function} column is specialized for 2 layers of MLP and GCN. $\textrm{SMx}(\cdot)$ refers to the softmax layer. For brevity, we use $S$ for both feature and label propagation matrices. The transformed feature matrix (${\bar X}$) can be pre-computed when the feature propagation component is the first component in the network (e.g., SGCN, FP + MLP and first layer in GCN). We assume that $S^2$ is split equally in FP and LP components (e.g., SGCN + LP, GCN + LP). In the \textit{cost} column, $L$ and $L_l$ denote the number of layers (in GCN or MLP) and in the label propagation network respectively. $n,N_E$ and $M$ denote the number of nodes, number of edges in $S$ and number of classes respectively.  
For simplicity, we assume that the feature dimension is fixed at $d$ in all layers.} 
\label{tab:std_split}
\begin{tabular}{|l|l|l|}
\hline
\textbf{Network}    & \textbf{Function} (2 layers) & \textbf{Cost} \\ \hline
\textbf{GCN}               & $\textrm{SMx}(\bm{S}\textrm{ReLU}({\bar X}W) W_S)$         & $O(LN_Ed + Lnd^2 + ndM)$ \\ \hline
\textbf{SGCN}        & $\textrm{SMx}({\bar X}W_S)$        & $O(ndM)$   \\ \hline
\textbf{FP + MLP} & $\textrm{SMx}(\textrm{ReLU}({\bar X}W)W_S)$         & $O(Lnd^2+ndM)$  \\ \hline
\textbf{SGCN + LP}            & $\bm{S} (\textrm{SMx}({\bar X}W_S))$         & $O(L_lN_EM + ndM)$  \\ \hline
\textbf{GCN+LP}             & $\bm{S} (\textrm{SMx}(\textrm{ReLU}({\bar X}W) W_S))$        & $O(LN_Ed + Lnd^2 + ndM + L_lN_EM)$   \\ \hline
\textbf{Linear + LP}    & $\bm{S}^2( \textrm{SMx}(XW_S))$          & $O(L_lN_EM+ndM)$  \\ \hline
\textbf{MLP+LP} & $\bm{S}^2\textrm{SMx}(\textrm{ReLU}(XW)W_S)$         & $O(Lnd^2+ L_lN_EM + ndM)$  \\ \hline

\end{tabular}
\end{table}

\section{Experiments}
We conducted a detailed experimental study to evaluate and compare performances of model variants on several benchmark datasets for different labeled data sizes. Due to space constraint, we give results only for small and large labeled data sizes here and other results can be found in the appendix. We start with providing details of datasets, baselines and training. Then, we present several key observations from our evaluation. 
\subsection{Experimental Setup}
\subsubsection{Datasets}
We evaluate on five datasets: Cora, Citeseer, Pubmed taken from~\cite{ica}, and ACM, DBLP taken from~\cite{han}. The datasets Cora, CiteSeer, Pubmed and ACM have scientific papers as nodes. In DBLP dataset, each node is an author. The node features in all these datasets correspond to sparse bag of words representation with binary or TF-IDF values. Class labels description for each dataset can be found in the appendix. We fix the validation and testing set to 500 and 1000 nodes respectively. All the nodes that do not belong to validation or test set contribute towards the training data; we refer to this set as T. The smallest training dataset is created by sampling 20 nodes per class from T to be in line with the standard split. Let M denote the number of classes in the dataset. We divide [20M,  $|\textrm{T}|$] into 3 equal intervals, which gives us 5 different training data sizes. We sample nodes from T and create training sets corresponding to these sizes. To ensure training consistency, we make the smaller training sets subset of the larger training sets.

The detailed statistics of the datasets are tabulated in Table 2. Node feature vectors were normalized to unit vectors and the augmented adjacency graph was symmetrically normalized \cite{gcn}. For label propagation variants, the above graph was row normalized to ensure probability outputs. In literature, many works report performance on a standard split of the above mentioned datasets. However, a single split offers limited information about generalization performance of the classification models. To address this, we evaluate on standard data split (for the purpose of reference) and 10 random splits (with standard data split size and $4$ different labeled dataset sizes). Our experimental results on the standard data split can be found in the appendix and they match the performances of existing models reported in the literature. We give results only for small and large labeled data sizes here and other results can be found in the appendix.

\begin{table}[]
\centering
\caption{Datasets Statistics}
\label{tab:cv_main_few}
\resizebox{\textwidth}{!}{%
\begin{tabular}{|c|c|c|c|c|c|}
\hline
\textbf{} & \textbf{Cora} & \textbf{Citeseer} & \textbf{Pubmed} & \textbf{ACM} & \textbf{DBLP} \\ \hline
\textbf{Nodes}         & 2,708 & 3,327 & 19,717 & 3,025 & 4,057 \\ \hline
\textbf{Edges}        & 5,278 & 4,552 & 44,324 & 16,153 & 7,585 \\ \hline
\textbf{Features} & 1,433 & 3,703 & 500 & 1,830 & 334 \\ \hline
\textbf{Classes}     & 7 & 6 & 3 & 3 & 4 \\ \hline
\textbf{Train size}      & [140, 1208] & [120, 1827] & [60, 18217] & [60, 1525] & [80, 2557] \\ \hline 
\textbf{Edge Type}    & citation-link & citation-link & citation-link & author-link & paper-link \\ \hline
\end{tabular}%
}
\end{table}


\subsubsection{Baselines} We compare performances of newly composed variants with popular GCN \cite{gcn} and SGCN \cite{sgcn} models. Since we have introduced label propagation in the GCN context, we also consider another baseline algorithm, Label Propagation with Neural Networks (LPNN) \cite{lpnn} which optimizes an objective function that matches label propagation induced class distributions with predictions from a features based neural network classifier model. In our experiments, we set the number of layers ($L$) in MLP, GCN and propagation networks (i.e., $L_l)$ to 2. More details on hyper-parameter search space, LPNN algorithm and its objective function can be found in the appendix.   

\subsubsection{Training Specifications}
All baselines and composed models were implemented and evaluated in an identical setup. Each model was trained for 500 epochs with an early stopping criterion of no improvement in validation accuracy for 25 epochs. We used Adam optimizer to minimize the standard cross entropy loss for all the models (except LPNN). Hyper-parameter search was done using \textit{optuna} \cite{optuna}. For each dataset split, we swept through 2000 configurations using TPE sampler in \textit{optuna}. We computed test accuracy for the model configuration that achieves the maximum validation accuracy.

\subsection{Results}
\subsubsection{Small Labeled Data} 
In Table~\ref{tab:mainresults_small}, we report mean accuracy performances on benchmark datasets in small label set size scenario. In the top and middle boxes, we have results from existing popular GCN and newly composed models. We see that the performances of all composed models are quite competitive to existing models and the performances of most models differ within $1-2\%$. 

\textbf{Average Rank based Comparison} To get a global view of performances of different models, we compare average rank ($R$) of models. The average rank $R_i$ of $i^{th}$ model is computed as: $R_i = \frac{1}{K} \sum_{k=1}^K R_{i,k}$ where $R_{i,k}$ is the rank of $i^{th}$ model in the $k^{th}$ dataset. Models with lower average rank values are better in terms of performances assessed across all datasets. The following observations can be made from Table~\ref{tab:mainresults_small}. 
\begin{itemize}
\item The top performance is achieved by GCN + LP providing evidence that label propagation can help GCN to get improvement performance. Indeed, adding LP improves the performance in DBLP by 1.7\%. Note that GCN + LP consists of one layer of feature and label smoothing each. One possible explanation for improved performance is that multi-layer feature propagation with GCN may cause features to become more non-discriminatory due to presence of noisy edges (i.e., connected nodes belonging to different classes) and this issue seems to be mitigated by mixing feature and label propagation networks.   
\item We see that GCN and MLP with LP get almost same average score. A similar observation holds with SGCN and Linear + LP. Together, these results suggest that label propagation is as good as feature propagation.
\item The average ranks of all linear models (i.e., SGCN (4.8), SGCN + LP (4.6), Linear + LP (4.9)) are consistently higher than all the MLP (3.6, 4.0) and GCN variants (2.4, 3.7), indicating that nonlinear models help to get improved performances. However, linear models are sufficient or better for datasets like Citeseer and we hypothesize that this is due to sparse and high dimensional nature of features. 
\item Composed LP models perform significantly better than the LPNN model. One possible explanation is that using label propagation directly in learning model parameters may be more effective than using it as a regularizer.   
\end{itemize}

\setlength{\tabcolsep}{2.5pt}

\begin{table}[]
\centering
\caption{\textbf{Mean test accuracy averaged over 10 splits with few labels. No. of training labels are 140 for Cora, 120 for Citeseer, 60 for Pubmed, 60 for ACM and 80 for DBLP.}}
\label{tab:cv_main_few}
\resizebox{\textwidth}{!}{%
\begin{tabular}{|c|c|c|c|c|c|c|}
\hline
\textbf{Methods} & \textbf{Cora} & \textbf{Citeseer} & \textbf{Pubmed} & \textbf{ACM} & \textbf{DBLP} & \textbf{R} \\ \hline
\textbf{GCN}       & 82.2 (1.1) & 68.8 (2.1) & 78.9 (1.8) & 89.7 (1.1) & 74.3 (1.5) & 3.7 \\ 
\textbf{SGCN}      & 81.6 (1.2) & 69.0 (1.6) & 78.1 (1.5) & 89.2 (1.0) & 74.6 (1.8) & 4.8 \\ \hline
\textbf{FP+MLP}    & 82.4 (0.8) & 67.0 (1.7) & 78.3 (2.0) & 89.0 (1.5) & 75.9 (1.2) & 4.0 \\ 
\textbf{SGCN+LP}   & 81.4 (1.1) & \textbf{69.7} (2.3) & 78.2 (1.9) & 88.8 (1.7) & 75.1 (1.5) & 4.6 \\ 
\textbf{GCN+LP}    & 82.1 (1.1) & 68.9 (1.5) & \textbf{79.2} (2.2) & 89.2 (1.1) & \textbf{76.0} (1.6) & \textbf{2.4} \\ 
\textbf{Linear+LP} & 82.0 (1.2) & 68.8 (2.1) & 78.7 (1.7) & 88.7 (1.1) & 74.8 (2.2) & 4.9 \\ 
\textbf{MLP+LP}    & \textbf{82.7} (1.1) & 68.1 (1.8) & 78.1 (1.4) & \textbf{90.3} (1.1) & 75.6 (2.0) & 3.6 \\ \hline
\textbf{LPNN}      & 77.8 (2.2) & 59.6 (4.3) & 69.8 (4.3) & 78.9 (4.7) & 64.6 (3.4) & 8.0 \\ \hline
\end{tabular}%
}
\label{tab:mainresults_small}
\end{table}
\subsubsection{Large Labeled Data} As in the case of small labeled data scenario, the performances of all composed models are quite competitive to existing SGCN and GCN. However, there are two important differences: (1) LPNN model's performance improved significantly with increase in labeled dataset size even achieving the best performance in the Pubmed dataset. (2) Interestingly, SGCN becomes the top performer suggesting that simple models are quite competitive when sufficiently large number of labeled examples are available, particularly, when the number of features is reasonably high, as we found in text datasets.

Overall, our experimental study suggests that there is no single model that performs significantly and uniformly better across all datasets. This suggests the proposed model variants are useful to experiment and validate along with existing SGCN and GCN models, as new model variants may provide performance improvement in several applications.    

\begin{table}[]
\centering
\caption{\textbf{Mean test accuracy averaged over 10 splits with many labels. No. of training labels are 1208 for Cora, 1827 for Citeseer, 18217 for Pubmed, 1525 for ACM and 2557 for DBLP.}}
\label{tab:cv_main_few}
\resizebox{\textwidth}{!}{%
\begin{tabular}{|c|c|c|c|c|c|c|}
\hline
\textbf{Methods} & \textbf{Cora} & \textbf{Citeseer} & \textbf{Pubmed} & \textbf{ACM} & \textbf{DBLP} & \textbf{R} \\ \hline
\textbf{GCN}       & 87.5 (1.1) & 75.8 (1.5) & 87.7 (1.2) & \textbf{92.2} (0.8) & 84.0 (0.7) & 4.2 \\ 
\textbf{SGCN}      & 87.9 (0.6) & \textbf{76.0} (1.3) & 86.4 (1.1) & \textbf{92.2} (0.9) & 84.2 (0.9) & \textbf{2.6} \\ \hline
\textbf{FP+MLP}    & 87.8 (1.0) & 75.5 (1.4) & 88.2 (1.0) & 92.1 (0.6) & 84.1 (0.5) & 3.8 \\ 
\textbf{SGCN+LP}   & 87.5 (0.9) & \textbf{76.0} (1.2) & 85.9 (1.4) & 91.9 (1.3) & 84.4 (0.8) & 4.6 \\ 
\textbf{GCN+LP}    & 87.7 (1.1) & 75.7 (1.4) & 87.9 (0.9) & 91.8 (1.0) & 83.9 (0.6) & 5.0 \\ 
\textbf{Linear+LP} & 87.5 (1.2) & 75.9 (1.5) & 85.3 (1.3) & 91.8 (1.1) & 83.8 (0.8) & 5.8 \\ 
\textbf{MLP+LP}    & \textbf{88.1} (0.7) & 75.6 (2.0) & 86.7 (1.2) & 92.0 (0.7) & \textbf{84.7} (0.7) & 3.4 \\ \hline
\textbf{LPNN}      & 86.0 (1.2) & 73.8 (1.6) & \textbf{90.1} (0.6) & 91.4 (1.4) & 83.0 (0.7) & 6.6 \\ \hline
\end{tabular}%
}
\end{table}
  
\subsubsection{Experiments with Propagation Models}
To study the effect of the number of layers ($L_l$) in label propagation, we conducted an experiment to compare performances achieved for different $L_l$ ($[1,4]$) on Cora and Citeseer datasets. We observed that that $L_l = 2$ or $3$ gives good performance and there is no significant improvement achieved by further increasing $L_l$. These results match the observations reported in \cite{sgcn} for feature propagation. More details of results obtained are given in the appendix. 

We also experimented with two label propagation models that are specified by hyper-parameters. The first model is specified as: ${\tilde A} = (\alpha I + \beta A)$ where $0 \le \alpha, \beta \le 1$ with $\alpha$ and $\beta$ controlling the degree at which self and neighboring nodes influence the class probability distribution. In a general setting, $\alpha$ and $\beta$ are $n$-dimensional vectors with elements controlling individual nodes. The second model is specified as: ${\tilde S} = {\tilde D}^{-\alpha}{\tilde A}{\tilde D}^{-\beta}$. Our preliminary experiments on Cora and Citeseer datasets show 1-2\% performance improvement with hyper-parameter tuning. A detailed experimental study with propagation models is left as future work.

\section{Related Work}
Graph-based semi-supervised learning approaches have been well-studied in the literature. Such approaches have seen improved performances even in the presence of very few labels. All these models make the underlying assumption that the neighbouring nodes in a graph are likely to have similar labels.

Early approaches cast the problem as a graph partitioning problem \cite{mincut}. However, there were two issues: (a) graph partitions are not unique with many solutions being degenerate, and (b) generalizing the idea to multi-class classification problem is expensive. Both these issues were addressed by using label propagation~\cite{lp_zhu,lp_zhu_zoubin_lafferty} and its variants~\cite{lp_zhou,lp_chapelle,adsorption}. Label propagation is a popular algorithmic approach in which labels of unlabeled nodes are inferred by propagating \textit{known labels} through neighbors (i.e., using proximity information available) of each node. In essence, a label propagation algorithm is iterative and estimates each node's label by aggregating labels of its neighbors and passes estimated label information to neighbors for further processing. 

The aforementioned works do not account for node features in their modeling. This gave rise to a closely related algorithmic approach - graph transduction \cite{manifold_reg,altmin_wang} where a classification function that (a) ensures smoothness over the graph, and (b) respects known label information is learned. In this approach, an objective function that comprises of one or more regularization terms is defined. These regularization terms penalize violation of constraints such as function smoothness and matching known labels.

Recent works have started looking at this problem in a new way by generalizing convolutional neural networks to graph data~\cite{graph_cnn,gcn}. These models define a reference operator for graph fourier transform, which is usually a symmetric normalized graph Laplacian. They then apply a filter on the spectral components of this reference operator. Graph Convolutional Network (GCN)~\cite{gcn} is the state-of-the-art in this series of works. It simplifies the model significantly by restricting the filter to operate on just 1-hop neighbourhood while retaining good performance. A closer look at GCN suggests that the model works by propagating \textit{features} from each node to its neighbours, instead of labels (as done in the earlier label propagation models). Simplified GCN (SGCN)~\cite{sgcn} questions the presence of non-linear layers in the GCN model, thereby, proposing a linearized version of GCN that retains good performance (with feature propagation) while giving significant training speedup.

Our proposed framework defines smoothing layer that can be used with labels and/or features, thus allowing us to view the label propagation and feature propagation models through a common lens. This allows us to ask and answer several questions including comparison of label and feature propagation, and effectiveness of using non-linearity. Additionally, the framework allows to construct new variants that perform as well as, and in some cases, better than state-of-the-art models.

Our work is closely related to EmbedNN\footnote{There is also a software framework, Deep Graph Library (DGL)~\cite{dgl}, which abstracts several graph based SSL models, however, the goals and purposes of DGL are different from ours.}~\cite{embednn} in which the authors propose a graph-based regularization term that can be applied to any layer or set of layers of a multi-layer neural network. The authors in~\cite{lpnn} noted that EmbedNN model is susceptible to poor performance due to noisy/missing features, and proposed a model that combines label propagation with neural networks to give improved performance. Our results seem to suggest that feature/label smoothing may have an advantage over regularization based approach.

It is also worthwhile to note that there are other feature propagation models that do not fit into our framework. For instance, GraphSAGE~\cite{hamilton2017inductive} which takes the idea of feature propagation and learns a mapping between representation of the neighbours of a node, and the node, thus allowing it to inductively predict representations of new nodes when they get added to the graph. Another model, Graph Attention Networks (GAT)~\cite{gat} learns an attention function over the edges of the graph, and features are propagated as per the learnt weights on the edges, thus avoiding any necessity for computation of normalized Laplacian, which could be costly. Embedding Propagation (EP)~\cite{embedding_propagation} is an unsupervised method which learns latent representations for the nodes of the graph by propagating representations of node attributes along the graph. 
\section{Discussion and Future Work}

In propagation networks, it is useful to incorporate learnable or model selection capabilities in propagation models (i.e., $S$) by introducing a few learnable parameters or hyper-parameters to get improved performance. We conducted preliminary experiments with two simple hyper-parameterized propagation models and observed 1-2\% accuracy improvement in Cora and Citeseer datasets. Designing hyper-parameterized propagation models and validating usefulness of such models are promising future works to carry out.    

One important area gaining popularity recently is to develop robust methods in the noisy graph scenario. This is because an assumption usually made in graph based SSL methods is that two nodes connected by an edge have same label. However, this assumption does not hold in practice and noisy edges are present in real-world datasets. Recently, a graph agreement modeling approach \cite{gam} was proposed, where an edge classifier model that predicts $+1$ for edges connecting nodes with same labels and $0$ otherwise is learned using node features. Then, the node classifier model (e.g., GCN) is learned such that its predictions are in agreement with the edge classifier model and this is achieved by defining a new loss function with regularization terms that penalize disagreement. Another future direction is to integrate agreement/regularization models with the models composed from our framework.

\section{Conclusion}
We studied the problem of designing variants of GCNs and proposed a framework to compose networks using the building blocks of GCN. We introduced propagation networks and showed how label propagation can be incorporated in a simple way with linear and nonlinear models including existing GCNs and SGCNs. Our experimental results show that label propagation is as good as feature propagation in several datasets. Furthermore, newly composed networks achieve competitive performances and label propagation can help to get improved performance with GCN and SGCN. Finally, with no single model achieving the best performance across all datasets, the framework offers a variety of GCN variants along with SGCN and GCN to experiment with and validate on real-world datasets.   

\bibliographystyle{extras/splncs04}
\bibliography{references}
\clearpage
\title{Appendix}

\titlerunning{GCN Composition Framework}

\author{}

\institute{}
\authorrunning{R. Ragesh et al.}

\maketitle 
\section{Extended Results}

\subsection{Effect of Increasing Training Labels}
We carry out a comprehensive evaluation of the baselines and our proposed variants on Cora, Citeseer, Pubmed, ACM and DBLP. To study the effect of number of training labels, we create five different sized labelled sets for each of the datasets. For each sized set, we further sample ten random splits. Results for Set 1 (few labels) and Set 5 (many labels) are presented in the main paper. The results for Set 2, Set 3 and Set 4 are presented in Tables \ref{tab:cv_results_2}, \ref{tab:cv_results_3} and \ref{tab:cv_results_4}  below.  
\vspace{-18pt}
{ 
    \setlength{\tabcolsep}{9pt}
    \begin{table}
\centering
\caption{\textbf{Mean test accuracy averaged over 10 splits (Set 2). No. of training labels are 407 for Cora, 547 for Citeseer, 4600 for Pubmed, 426 for ACM and 699 for DBLP.}}
\label{tab:cv_results_2}
\resizebox{\textwidth}{!}{%
\begin{tabular}{|c|c|c|c|c|c|c|}
\hline
\textbf{Methods} & \textbf{Cora} & \textbf{Citeseer} & \textbf{Pubmed} & \textbf{ACM} & \textbf{DBLP} & \textbf{R} \\ \hline
\textbf{GCN}       & 85.7 (0.8) & 72.8 (0.9) & 86.6 (1.2) & 91.6 (1.3) & 81.7 (1.3) & 3.8 \\ 
\textbf{SGCN}      & 85.0 (0.8) & 73.1 (1.1) & 86.5 (0.9) & 91.8 (1.2) & 81.7 (1.1) & 4.7 \\ \hline
\textbf{FP+MLP}    & 85.3 (0.9) & 72.4 (1.1) & 86.5 (1.1) & 91.3 (1.1) & 82.6 (0.9) & 4.5 \\ 
\textbf{SGCN+LP}   & 85.1 (0.7) & 73.3 (0.8) & 86.0 (1.2) & 91.3 (1.1) & 82.1 (0.7) & 4.7 \\ 
\textbf{GCN+LP}    & 85.0 (0.8) & 73.2 (1.0) & 86.8 (1.4) & 91.4 (1.3) & 82.4 (1.1) & 3.5 \\ 
\textbf{Linear+LP} & 85.3 (0.8) & 73.1 (0.9) & 85.1 (1.2) & 91.4 (1.0) & 82.0 (1.1) & 5.2 \\ 
\textbf{MLP+LP}    & 85.5 (0.9) & 72.8 (1.4) & 86.2 (0.8) & 91.9 (1.1) & 82.6 (1.0) & 3.0 \\ \hline
\textbf{LPNN}      & 82.5 (1.1) & 67.7 (1.8) & 87.7 (1.0) & 89.5 (1.9) & 78.2 (1.3) & 6.6 \\ \hline
\end{tabular}%
}
\end{table}}
    
\vspace{-18pt}
{ 
    \setlength{\tabcolsep}{9pt}
    \begin{table}
\centering
\caption{\textbf{Mean test accuracy averaged over 10 splits (Set 3). No. of training labels are 674 for Cora, 974 for Citeseer, 9139 for Pubmed, 792 for ACM and 1318 for DBLP.}}
\label{tab:cv_results_3}
\resizebox{\textwidth}{!}{%
\begin{tabular}{|c|c|c|c|c|c|c|}
\hline
\textbf{Methods} & \textbf{Cora} & \textbf{Citeseer} & \textbf{Pubmed} & \textbf{ACM} & \textbf{DBLP} & \textbf{R} \\ \hline
\textbf{GCN}       & 86.3 (0.9) & 74.5 (1.3) & 87.3 (1.1) & 92.4 (0.9) & 82.7 (1.2) & 4.0 \\ 
\textbf{SGCN}      & 86.1 (1.0) & 74.7 (1.4) & 86.3 (1.0) & 92.0 (0.8) & 83.2 (0.8) & 4.8 \\ \hline
\textbf{FP+MLP}    & 86.5 (0.8) & 74.1 (1.5) & 87.1 (1.1) & 91.6 (1.0) & 83.6 (0.7) & 4.2 \\ 
\textbf{SGCN+LP}   & 84.0 (6.3) & 74.6 (1.4) & 85.9 (1.8) & 92.0 (1.0) & 83.3 (1.0) & 4.6 \\ 
\textbf{GCN+LP}    & 86.1 (1.2) & 74.9 (1.2) & 87.3 (1.3) & 92.0 (0.9) & 83.1 (0.8) & 3.4 \\ 
\textbf{Linear+LP} & 86.3 (1.3) & 74.8 (1.1) & 85.5 (1.3) & 91.7 (1.1) & 83.3 (0.9) & 4.4 \\ 
\textbf{MLP+LP}    & 86.7 (1.0) & 74.0 (1.2) & 86.3 (1.5) & 91.7 (0.7) & 83.8 (0.6) & 4.0 \\ \hline
\textbf{LPNN}      & 83.7 (0.8) & 70.8 (1.2) & 89.0 (1.0) & 90.3 (0.9) & 80.0 (1.5) & 6.6 \\ \hline
\end{tabular}%
}
\end{table}}

\vspace{-18pt}
{ 
    \setlength{\tabcolsep}{9pt}
    \begin{table}
\centering
\caption{\textbf{Mean test accuracy averaged over 10 splits (Set 4). No. of training labels are 941 for Cora, 1401 for Citeseer, 13678 for Pubmed, 1158 for ACM and 1937 for DBLP.}}
\label{tab:cv_results_4}
\resizebox{\textwidth}{!}{%
\begin{tabular}{|c|c|c|c|c|c|c|}
\hline
\textbf{Methods} & \textbf{Cora} & \textbf{Citeseer} & \textbf{Pubmed} & \textbf{ACM} & \textbf{DBLP} & \textbf{R} \\ \hline
\textbf{GCN}       & 87.2 (0.9) & 75.5 (1.3) & 87.6 (1.2) & 92.4 (0.6) & 83.5 (1.1) & 4.0 \\ 
\textbf{SGCN}      & 87.2 (1.1) & 75.5 (1.6) & 86.2 (1.1) & 92.4 (1.1) & 83.6 (1.3) & 3.4 \\ \hline
\textbf{FP+MLP}    & 87.5 (0.8) & 75.0 (1.5) & 88.1 (1.0) & 91.8 (1.0) & 84.1 (0.8) & 3.6 \\ 
\textbf{SGCN+LP}   & 87.7 (0.8) & 75.3 (1.5) & 85.9 (1.2) & 91.8 (0.5) & 83.6 (0.8) & 4.9 \\ 
\textbf{GCN+LP}    & 86.9 (1.0) & 75.8 (1.4) & 87.3 (1.1) & 91.9 (0.8) & 83.2 (1.3) & 4.4 \\ 
\textbf{Linear+LP} & 86.7 (1.3) & 75.9 (1.4) & 85.4 (1.3) & 91.6 (0.9) & 83.6 (0.8) & 5.5 \\ 
\textbf{MLP+LP}    & 87.7 (0.7) & 75.4 (1.2) & 86.7 (1.6) & 91.9 (0.7) & 84.1 (0.5) & 3.6 \\ \hline
\textbf{LPNN}      & 85.0 (0.6) & 72.1 (1.4) & 89.7 (0.7) & 90.8 (1.4) & 81.4 (0.9) & 6.6 \\ \hline
\end{tabular}%
}
\end{table}}

\subsection{Results on Standard Split}
Most works in the literature evaluate the models on the standard splits of Cora, Citeseer and Pubmed. We also report numbers on these splits in Table \ref{std_main_few}. It is interesting to note that from our experiments, with thorough tuning, GCN numbers are 1-2 \% higher than the reported numbers on Cora and Citeseer in literature. 

{ 
    \setlength{\tabcolsep}{9pt}
    \begin{table}
    \centering
    \caption{\textbf{Test accuracy on the standard split.}}
    \label{std_main_few}
    \resizebox{0.75\textwidth}{!}{%
    \begin{tabular}{|c|c|c|c|c|}
    \hline
    \textbf{Methods}     & \textbf{Cora} & \textbf{Citeseer} & \textbf{Pubmed} & \textbf{R} \\ \hline
    \textbf{GCN}         & 83.5          & 72.3              & 79.3            & 2.7               \\
    \textbf{SGCN}        & 81.2          & 73.1              & 79.8            & 2.8               \\ \hline
    \textbf{FP+MLP}      & 82.0          & 69.0              & 78.1            & 5.3               \\
    \textbf{SGCN+LP}     & 81.9          & 71.7              & 79.0            & 4.3               \\
    \textbf{GCN+LP}      & 81.2          & 71.1              & 79.7            & 4.5               \\
    \textbf{Linear+LP}   & 79.0          & 73.1              & 79.8            & 3.3               \\
    \textbf{MLP+LP}      & 82.8          & 70.5              & 78.0            & 5.0               \\ \hline
    \textbf{LPNN}        & 76.1          & 55.2              & 74.0            & 8.0               \\ \hline
    \end{tabular}%
    }
    \end{table}
}

\section{Datasets}
We give a short description of the classes for each of the dataets.\\
\indent\textbf{Cora.} Each node can be classified into one of the following seven classes: \textit{Case Based}, \textit{Genetic Algorithms}, \textit{Neural Networks}, \textit{Probabilistic Methods}, \textit{Reinforcement Learning}, \textit{Rule Learning}, and \textit{Theory}.\\ 
\indent\textbf{Citeseer.} Each node can be classified into one of the following six classes: \textit{Agents}, \textit{Artificial Intelligence}, \textit{Databases}, \textit{Information Retrieval}, \textit{Machine Learning}, and \textit{Human Computer Interaction}.\\
\indent\textbf{PubMed.} Each node can be classified into one of the following three classes: \textit{Diabetes Mellitus, Experimental}, \textit{Diabetes Mellitus Type 1}, \textit{Diabetes Mellitus Type 2}.\\
\indent\textbf{ACM.} Each node can be classified into one of the following three classes: \textit{Database}, \textit{Wireless Communication}, and \textit{Data Mining}.\\
\indent\textbf{DBLP.} Each node can be classified into one of the following four classes: \textit{database}, \textit{data mining}, \textit{machine learning}, and \textit{information retrieval}.

\section{Experiments with Propagation Models}
\subsection{Effect of k / \# of layers}
To study the effect of the number of layers ($L_l$) in label propagation, we conducted an experiment to compare performances achieved for different $L_l$ ($[1,4]$) on Cora and Citeseer datasets. We observed that that $L_l = 2$ or $3$ gives good performance and there is no significant improvement achieved by further increasing $L_l$. These results match the observations reported in SGCN for feature propagation

{ 
    \setlength{\tabcolsep}{9pt}
\begin{figure}
\centering
 \captionsetup{type=table} 
\caption{\textbf{Effect of \# of layers / k }}
\begin{minipage}{\textwidth}
  \centering
  \small
  \captionsetup{type=table} 
  \caption*{\textbf{Citeseer}}
  \vspace{-9pt}
  \begin{tabular}{|c|c|c|c|c|}
\hline
\textbf{Methods\textbackslash{}k} & \textbf{1} & \textbf{2} & \textbf{3} & \textbf{4} \\ \hline
\textbf{SGCN}                     & 71.6       & 73.1       & 72.9       & 74.6       \\ \hline
\textbf{GCN}                      & 71.6       & 72.3       & 71.6       & 69.1       \\ \hline
\textbf{Linear+LP}                & 71.1       & 73.1       & 73.1       & 72.8       \\ \hline
\textbf{MLP+LP}                   & 71.3       & 70.5       & 71.3       & 72.2       \\ \hline
\end{tabular}
\end{minipage}
\\
\vspace{2pt}
\begin{minipage}{\textwidth}
  \centering
  \small
  \captionsetup{type=table} 
  \caption*{\textbf{Cora}}
  \vspace{-9pt}
\begin{tabular}{|c|c|c|c|c|}
\hline
\textbf{Methods\textbackslash{}k} & \textbf{1} & \textbf{2} & \textbf{3} & \textbf{4} \\ \hline
\textbf{SGCN}                     & 78.7       & 81.2       & 82.6       & 79.3       \\ \hline
\textbf{GCN}                      & 78.7       & 83.5       & 82.4       & 82.3       \\ \hline
\textbf{Linear+LP}                & 76.5       & 79.0       & 82.7       & 82.5       \\ \hline
\textbf{MLP+LP}                   & 78.7       & 82.8       & 82.3       & 82.4       \\ \hline
\end{tabular}
\end{minipage}
\vspace{-12pt}
\end{figure}
}

\subsection{Effect of Different Propagation Models}
We also experimented with two label propagation models that are specified by hyper-parameters. The first model is specified as: ${\tilde A} = (\alpha I + \beta A)$ where $0 \le \alpha, \beta \le 1$ with $\alpha$ and $\beta$ controlling the degree at which self and neighboring nodes influence the class probability distribution. In a general setting, $\alpha$ and $\beta$ are $n$-dimensional vectors with elements controlling individual nodes. The second model is specified as: ${\tilde S} = {\tilde D}^{-\alpha}{\tilde A}{\tilde D}^{-\beta}$. Our preliminary experiments on cora and citeseer datasets show 1-2\% performance improvement with hyperparameter tuning. A detailed experimental study with propagation models is left as future work.     
{ 
    \setlength{\tabcolsep}{9pt}
\begin{figure}
\centering
 \captionsetup{type=table} 
\caption{\textbf{MLP+LP with Propagation model}  \boldsymbol{${\tilde A} = (\alpha I + \beta A)$}}
\begin{minipage}{\textwidth}
  \centering
  \small
  \captionsetup{type=table} 
  \caption*{\textbf{Citeseer}}
  \vspace{-9pt}
  \resizebox{\textwidth}{!}{%
\begin{tabular}{|c|c|c|c|c|c|c|c|c|c|c|}
\hline
\boldsymbol{$\alpha$}         & 0     & 0.1   & 0.25  & 0.33  & 0.5   & 0.67  & 0.75  & 0.9   & 1    & 1     \\ \hline
\boldsymbol{$\beta$}          & 1     & 0.9   & 0.75  & 0.67  & 0.5   & 0.33  & 0.25  & 0.1   & 0    & 1     \\ \hline
\textbf{Val}  & 73.0 & 75.6 & 75.6 & 75.0 & 74.6 & 75.0 & 76.0 & 74.2 & 60.0 & 74.8 \\ \hline
\textbf{Test} & 70.6 & 70.7 & 71.8 & 72.3 & 71.2 & 70.4 & 70.6 & 71.9 & 55.0 & 72.5 \\ \hline
\end{tabular}%
}
\end{minipage}
\\
\vspace{2pt}
\begin{minipage}{\textwidth}
  \centering
  \small
  \captionsetup{type=table} 
  \caption*{\textbf{Cora}}
  \vspace{-9pt}
\resizebox{\textwidth}{!}{%
\begin{tabular}{|c|c|c|c|c|c|c|c|c|c|c|}
\hline
\boldsymbol{$\alpha$}         & 0     & 0.1   & 0.25  & 0.33  & 0.5   & 0.67  & 0.75  & 0.9   & 1     & 1     \\ \hline
\boldsymbol{$\beta$}          & 1     & 0.9   & 0.75  & 0.67  & 0.5   & 0.33  & 0.25  & 0.1   & 0     & 1     \\ \hline
\textbf{Val}  & 83.0 & 83.8 & 83.4 & 83.8 & 83.6 & 83.6 & 83.2 & 83.8 & 64.6 & 83.8 \\ \hline
\textbf{Test} & 81.7 & 82.3 & 81.6 & 83.2 & 83.4 & 82.7 & 82.3 & 81.8 & 59.0  & 81.6 \\ \hline
\end{tabular}%
}
\end{minipage}
\vspace{-12pt}
\end{figure}
}

{ 
    \setlength{\tabcolsep}{9pt}
\begin{figure}
\centering
 \captionsetup{type=table} 
\caption{\textbf{MLP+LP with Propagation model}  \boldsymbol{${\tilde D}^{-\alpha}{\tilde A}{\tilde D}^{-\beta}$}}
\begin{minipage}{\textwidth}
  \centering
  \small
  \captionsetup{type=table} 
  \caption*{\textbf{Citeseer}}
  \vspace{-9pt}
  \resizebox{\textwidth}{!}{%
    \begin{tabular}{|c|c|c|c|c|c|c|c|c|c|c|}
        \hline
        \boldsymbol{$\alpha$}         & 0     & 0.1   & 0.25  & 0.33  & 0.5   & 0.67  & 0.75  & 0.9   & 1     & 1     \\ \hline
        \boldsymbol{$\beta$}          & 1     & 0.9   & 0.75  & 0.67  & 0.5   & 0.33  & 0.25  & 0.1   & 0     & 1     \\ \hline
        \textbf{Val}  & 75.0  & 75.4 & 75.0  & 75.0  & 76.0 & 75.0  & 75.6 & 74.6 & 74.8 & 74.8 \\ \hline
        \textbf{Test} & 70.9 & 71.6 &  71.1 &  70.6 &  72.0 & 71.8 & 70.7 & 71.5 & 69.7 & 72.0  \\ \hline
    \end{tabular}%
    }
\end{minipage}
\\
\vspace{2pt}
\begin{minipage}{\textwidth}
  \centering
  \small
  \captionsetup{type=table} 
  \caption*{\textbf{Cora}}
  \vspace{-9pt}
\resizebox{\textwidth}{!}{%
\begin{tabular}{|c|c|c|c|c|c|c|c|c|c|c|}
\hline
\boldsymbol{$\alpha$}         & 0     & 0.1   & 0.25  & 0.33  & 0.5   & 0.67  & 0.75  & 0.9   & 1     & 1     \\ \hline
\boldsymbol{$\beta$}          & 1     & 0.9   & 0.75  & 0.67  & 0.5   & 0.33  & 0.25  & 0.1   & 0     & 1     \\ \hline
\textbf{Val}  & 84.4 & 83.8 & 83.8 & 84.0 & 84.0 & 83.6 & 83.4 & 83.6 & 83.8 & 83.6 \\ \hline
\textbf{Test} & 83.3 & 83.8 & 81.6 & 82.5 & 82.4 & 81.7 & 80.7 & 81.9 & 82.8 & 83.0  \\ \hline
\end{tabular}%
}
\end{minipage}
\vspace{-12pt}
\end{figure}
}

{ 
    \setlength{\tabcolsep}{8pt}
    \begin{table}
        \centering
        \caption{Hyperparameter Specifications}
        \label{tab:hyp_table}
        \resizebox{0.9\textwidth}{!}{
            \begin{tabular}{|c|c|c|}
                \hline
                \textbf{Hyperparamters} & \textbf{Search Space} & \textbf{Models} \\ \hline
            
                \begin{tabular}[c]{@{}c@{}}
                    learning rate, dropout, \\ weight decay\end{tabular} & uniform in (0,1) &  All \\ 
            
                \hline
            
                hidden dimensions &  [8, 16, 32, 64, 128] &  \begin{tabular}[c]{@{}c@{}}MLP based models, \\ GCN, LPGCN\end{tabular} \\ \hline
            
                \begin{tabular}[c]{@{}c@{}}$\mu_{G}$, $\mu_{l}$, $\mu_{u}$, \\ $\lambda_{l}$, $\lambda_{u}$\end{tabular} &  uniform in (0,1) &  LPNN \\ \hline
            \end{tabular}%
        }
    \end{table}
}
\section{Label Propagation with Neural Networks (LPNN) - Our Implementation}
LPNN encodes local continuity through $f$  while also utilizing the node features through $g$. It is worthwhile to note that our label propagation variants aggregates the probabilities, while in LPNN, label propagation is loosely used as a regularizer. Our implementation of LPNN jointly optimizes the following loss formulation \ref{eq:lpnn}, 

\begin{multline}
    \label{eq:lpnn}
    \mu_{G} Tr\{f^{T}(I_{N}-S)f\} + \mu_{l} \sum_{i=1}^{l}\norm{f_{i}-y_{i}}^{2} + \mu_{u}\sum_{i=l+1}^{n}\norm{f_{i}}^{2} + 
    \lambda_{l} \sum_{i=1}^{l} \textrm{KL}(y_{i}||g_{i})\\ + \lambda_{u} \sum_{i=l+1}^{n} \textrm{KL}(f_{i}||g_{i})
\end{multline}  

For LPNN, $h^b(.)$ and $g(f,x)$ consists of 2 hidden layers of dimensions 128 and 64 as specified in the original paper. We also set $h^a(.)$ to an identity function.

\subsection{Hyper-parameter Settings}
In table \ref{tab:hyp_table}, we specify the hyperparameters tuned for each of the model and the corresponding search space used in Optuna.

\end{document}